\documentclass[letterpaper, 10pt, conference]{ieeeconf}      % Use this line for a4 paper
\IEEEoverridecommandlockouts                              % This command is only needed if 
\usepackage{subfigure}
\usepackage{svg}
\usepackage{booktabs}
\usepackage{balance}
\usepackage{color}
\usepackage{tabularx}
\usepackage{graphics}
\usepackage{graphicx,dblfloatfix} 
\usepackage{epsfig} 
\usepackage{amsmath}
\usepackage{amssymb}
\usepackage{listings}
\usepackage{textcomp}
\usepackage{url}
\usepackage{float}
\usepackage{multirow}
\usepackage{todonotes}
\usepackage{hyperref}
\usepackage{cleveref}
\usepackage{wrapfig}
\crefformat{footnote}{#2\footnotemark[#1]#3}
\usepackage{ragged2e} 
\usepackage[subnum]{cases}
\newcolumntype{Y}{>{\RaggedRight\arraybackslash}X} 
\usepackage{algorithm}
\usepackage[noend]{algpseudocode}

\usepackage{tikz}
\usepackage{tkz-euclide}
\usetikzlibrary{math}
\usepackage{pgfplots}
\usetikzlibrary{arrows}

\usepackage{multicol, blindtext}

\pgfplotsset{compat=1.15}

\newcommand{\comment}[1]{}

\newcommand{\edited}[1]{\textcolor{black}{#1}}

\title{\LARGE \bf Bruce - Design and Development of a Dynamic Hexapod Robot*\thanks{* This research was, in part funded by the US Government under the DARPA Subterranean Challenge. The views, opinions, and findings expressed are those of the authors and should not be interpreted as representing the official views or policies of the Department of Defense or the U.S. Government. Approved for Public Release, Distribution Unlimited.}}

\author{Ryan Steindl, Thomas Molnar, Fletcher Talbot, Nicolas Hudson, Benjamin Tam, Simon Murrell, Navinda Kottege % Author list TBD  % <-this % stops a space
\thanks{R. Steindl, T. Molnar, F. Talbot, N. Hudson, B. Tam, S. Murrell and N. Kottege are with the Robotics and Autonomous Systems Group, CSIRO, Pullenvale, QLD 4069, Australia. All correspondence should be addressed to {\tt\small navinda.kottege@csiro.au}}%
}

\begin{document}
\maketitle
\thispagestyle{empty}
\pagestyle{empty}

%%%%%%%%%%%%%%%%%%%%%%%%%%%%%%%%%%%%%%%%%%%%%%%%%%%%%%%%%%%%%%%%%%%%%%%%%%%%%%%%%%%%%%%%%%%%%%%%%%%
%%%%%%%%%%%%%%%%%%%%%%%%%%%%%%%%%%%%%%%%%%%%%%%%%%%%%%%%%%%%%%%%%%%%%%%%%%%%%%%%%%%%%%%%%%%%%%%%%%%

\begin{abstract}
This paper introduces Bruce, the CSIRO Dynamic Hexapod Robot capable of autonomous, dynamic locomotion over difficult terrain. This robot is built around Apptronik linear series elastic actuators, and went from design to deployment in under a year by using approximately 80\% 3D printed structural (joints and link) parts. The robot has so far demonstrated rough terrain traversal over grass, rocks and rubble at 0.3\,ms$^{-1}$, and flat-ground speeds up to 0.5\,ms$^{-1}$. This was achieved with a simple controller, inspired by RHex, with a central pattern generator, task-frame impedance control for individual legs and no foot contact detection. The robot is designed to move at up to 1.0\,ms$^{-1}$ on flat ground with appropriate control, and  was deployed into the the DARPA SubT Challenge Tunnel circuit event in August 2019.
\end{abstract}

\section{Introduction}
\label{sec:introduction}
Multi-legged robots have the advantage over wheeled and tracked counterparts when traversing rough and unstructured terrain. Subterranean environments such as man-made tunnels, underground urban infrastructure and natural caves are examples of environments where the versatility of legged systems can be exploited for gaining rapid situational awareness through exploration. Existing robot research platforms have focused on quadruped locomotion, such as ANYmal~\cite{hutter_anymal_2016}, HyQ2Max~\cite{semini_design_2017}, TITAN-XIII~\cite{Kitano_2013} and MIT Cheetah\,3~\cite{bledt_mit_2018}. The design focus of quadruped robots have been for fast, dynamic locomotion (primarily on flat terrain) with increasing robustness against disturbances and uneven terrain. On the other hand, hexapod robots such as LAURON V~\cite{roennau_lauron_2014}, MAX~\cite{elfes_multilegged_2017}, RHex~\cite{saranli_rhex} and Snake Monster~\cite{travers_dynamical_2016} have focused on rough terrain locomotion as they are inherently more stable over rough terrain with a wider support polygon, lower centre of gravity, larger locomotion workspace and statically stable, fast gaits. Hexapod robots trade-off stability and speed in rough, unstable and slippery terrain for additional hardware complexity and weight compared to quadruped robots. In reality, environments contain a mix of flat and rough terrains, thus a robot platform capable of fast dynamic locomotion on flat terrain and fast stable locomotion on rough terrain would be ideal for time critical applications.

\begin{figure}[t!]
\centering
\includegraphics[width=7cm]{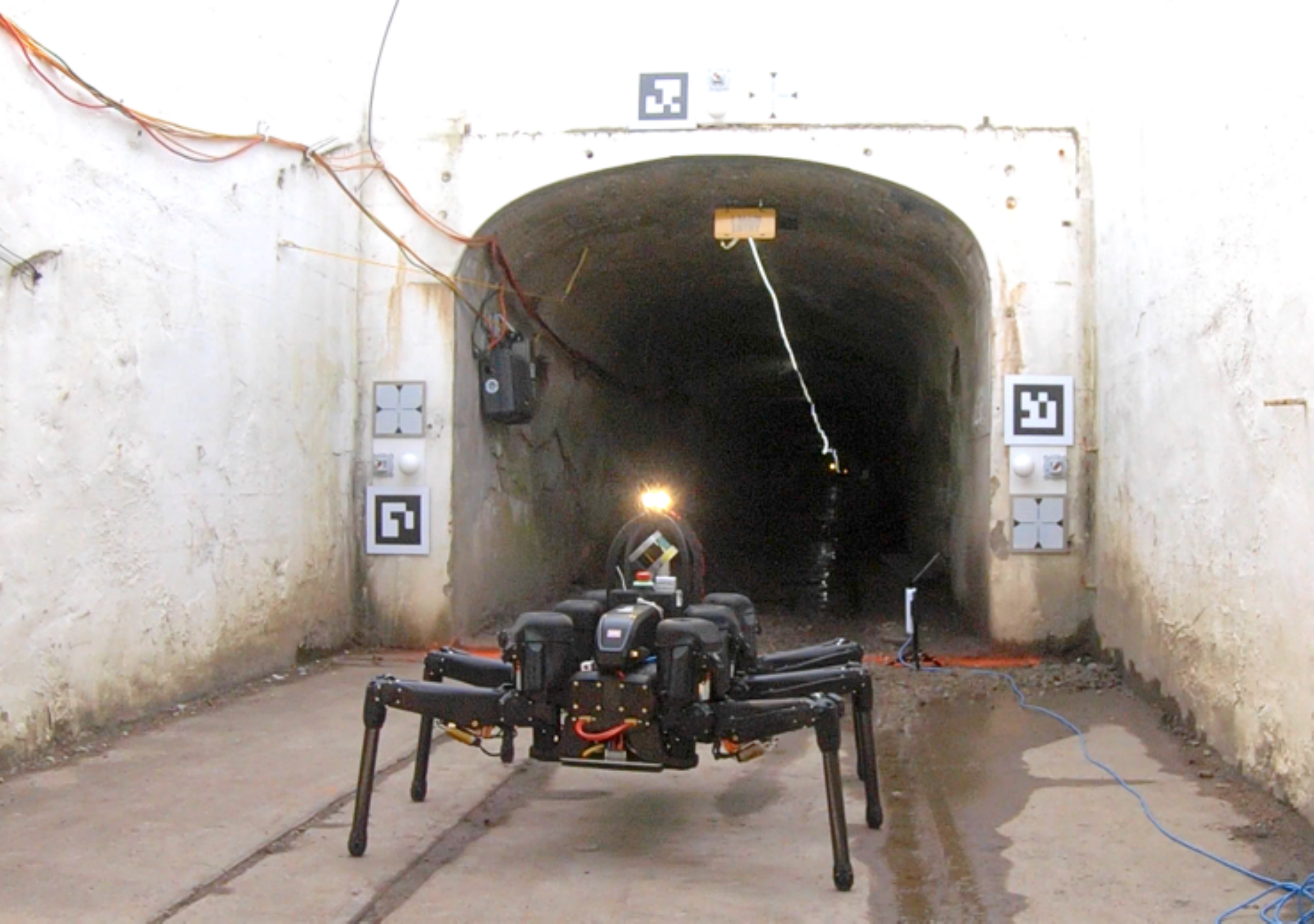}
\caption{The CSIRO Dynamic Hexapod Bruce, at the staging area before deployment into the Experimental Research mine during the DARPA SubT Challenge Tunnel Circuit event.}	
\label{fig:Bruce_at_tunnel}
 %%\vspace{-0.5cm}
\end{figure}

The authors present a novel hexapod robot platform called \textit{Bruce} (Figure~\ref{fig:Bruce_at_tunnel}), which was specifically designed to meet this paradigm. Bruce was originally intended to compete in the DARPA Subterranean Challenge, wherein it would be required to rapidly and autonomously explore an unknown subterranean environment composed of man-made and natural terrain formations. This paper focuses on the unique design using linear series elastic actuators (SEAs) \cite{paine_high-performance_2014} and the control strategy employed to achieve robust locomotion. 
\edited
{
The rest of the paper presents the design philosophy and system overview of both hardware design and control strategy used in Bruce. The design overview is given in Section~\ref{sec:designoverview}. Section~\ref{sec:hardwaredesign} presents the hardware design of the robot while the software components are described in Section~\ref{sec:software}. The robot's performance is evaluated in Section~\ref{sec:evaluation} and Section~\ref{sec:conclusions} concludes the paper.
}

\section{Design overview}
\label{sec:designoverview}
A hexapod robot design allows for its centre of mass (CoM) to be within its large support polygon while using the statically stable, fast tripod gait. This gait is also favoured by insects when traversing 3D terrain \cite{ramdya_climbing_2017}. However, compared to quadruped robots, where multiple platforms are able to achieve speeds greater than 1\,ms$^{-1}$ \cite{hutter_anymal_2016,ghostrobotics,Kitano_2013,bledt_mit_2018,unitree,bd_youtube_2018,semini_design_2017}, hexapod robots have not been able to travel at this speed, with the exception of RHex \cite{campbell_preliminary_2003,holmes_dynamics_2006}, a six degree of freedom (DOF) hexapod robot with compliant legs. 

Bruce was designed to have: most of its mass in the single central body, very light legs, and high bandwidth task frame impedance for each leg. This enables the system to respond similar to the corresponding spring-loaded inverted pendulum (SLIP) model, which has been shown to be a good representation for the dynamics of most legged robots and animals, including hexapods \cite{holmes_dynamics_2006}. Thus, mimicking hexapods in nature, Bruce achieves locomotion through tuned leg impedance feedback combined with an open loop gait engine akin to a central pattern generator (CPG). This approach was shown on RHex \cite{saranli_rhex} to lead to dynamic stability at higher speeds and allow traversing rough terrain without explicit detection of foot-ground contact. Bruce expands this idea to a three degree of freedom leg, requiring both passive and active software controlled impedance control to gain compliance. 

\begin{figure}[t!]
\centering
\includegraphics[width=8.5cm]{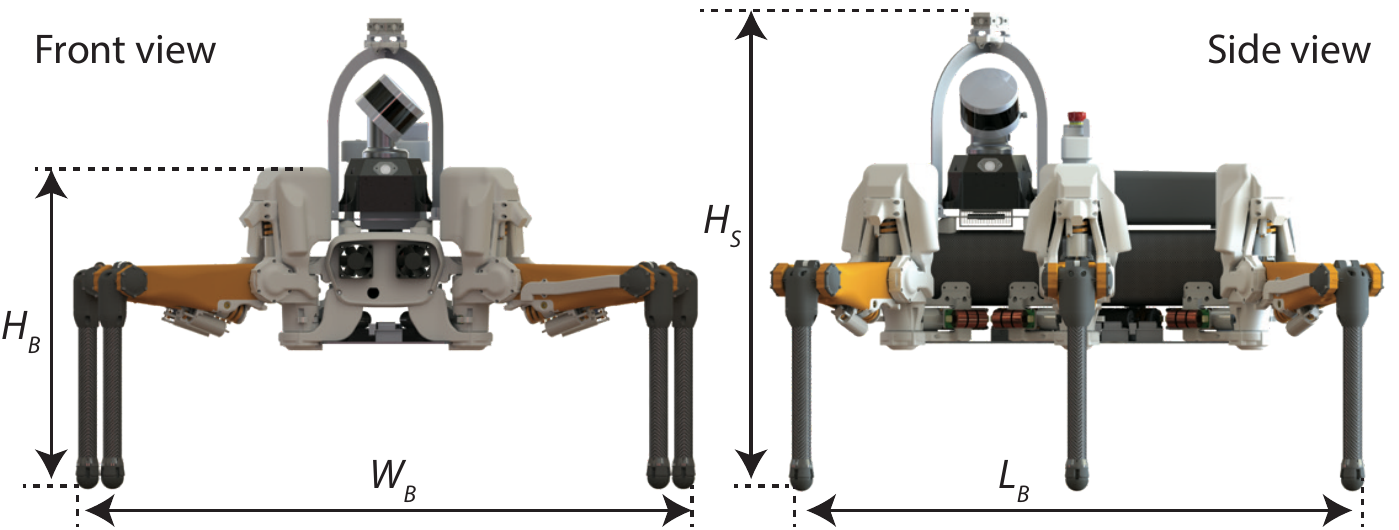}
\caption{Dimensions of Bruce when in a nominal stance.}	
\label{fig:dsh1dimensions}
\vspace{-0.5cm}
\end{figure}
%%%%%%%%%%%%%%%%%%%%%%%%%%%%%%%%%%%%%%%%%%%%%%%%%%%%%%%%%%%%%%%%%%%%%%%%%%%%%%%%%%%%%%%%%%%%%%%%%%%
%%%%%%%%%%%%%%%%%%%%%%%%%%%%%%%%%%%%%%%%%%%%%%%%%%%%%%%%%%%%%%%%%%%%%%%%%%%%%%%%%%%%%%%%%%%%%%%%%%%
%%%%%%%%%%%%%%%%%%%%%%%%%%%%%%%%%%%%%%%%%%%%%%%%%%%%%%%%%%%%%%%%%%%%%%%%%%%%%%%%%%%%%%%%%%%%%%%%%%%
\begin{figure}[b!]
\vspace{-0.5cm}
\centering
\includegraphics[width=5cm]{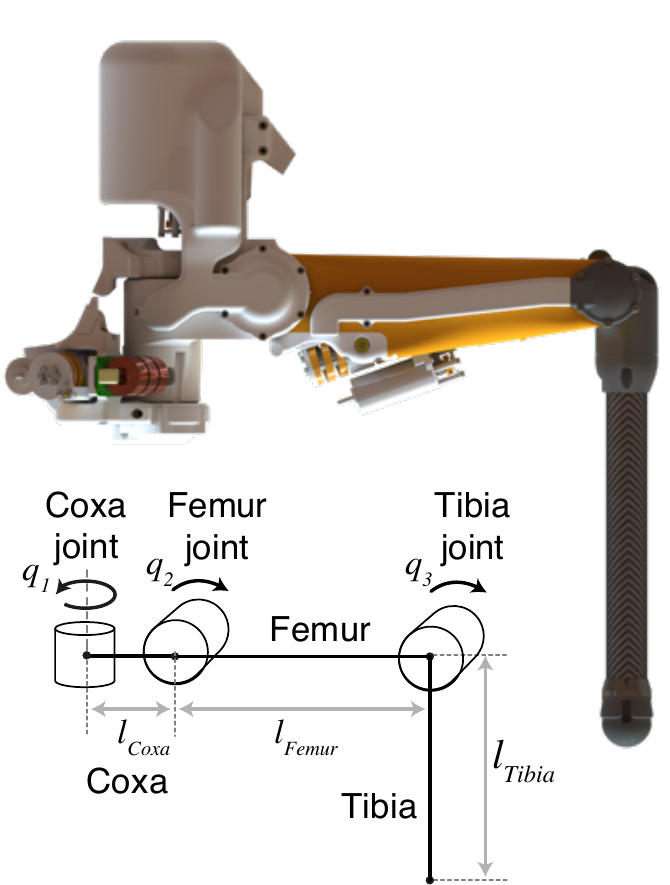}
\caption{Leg configuration of Bruce.}
\label{fig:legdimensions}
\end{figure}

\section{Hardware Design}
\label{sec:hardwaredesign}

\subsection{Morphology Requirements}
The performance requirements for the robot to be successful in the SubT Challenge governed many aspects of the robot's morphology. Rectilinear configuration of the legs was chosen to increase the available forward velocity by expanding the foot tip workspaces in the forward-aft direction. The total body length was limited to allow for reconfiguration of its footprint to strafe through 0.8\,m openings or stairwells. This resulted in the pivot points of the coxa joints being placed 0.3\,m apart and the foot tip workspaces to be 0.5\,m long, giving a 0.1\,m safety buffer to avoid leg collisions. Link lengths, and required joint velocities and torques were optimised in simulation to determine the final dimensions of the legs that achieved the target body velocity of 1.0\,ms$^{-1}$. The dimensions of the robot are shown in Figure~\ref{fig:dsh1dimensions} and Figure~\ref{fig:legdimensions}, with all technical specifications outlined in Table~\ref{table:specs}.

\begin{table}[b!]
    \caption{Hardware Specifications of Bruce.}  
    \label{table:specs}
    \vspace{-3.5ex}
    \begin{center}
    \begin{tabularx}{3.3 in}{@{} c Y @{}}
    %\begin{tabular}{cl}
    \toprule
    Type & Description \\
    \midrule
    General & Dimensions (stance): $L_{B}$ = 1.0\,m $\times$ $W_{B}$ = 1.08\,m $\times$ $H_{B}$ = 0.55\,m ($H_{S}$ = 0.8\,m)\\
            & Mass (without battery): 47.18\,kg\\ 
            & Mass (with battery): 50.5\,kg\\
            & Mass (payload): 3.2\,kg\\
    Leg & Dimensions: $l_{Coxa}$ = 0.065\,m $\times$ $l_{Femur}$ = 0.275\,m $\times$ $l_{Tibia}$ = 0.365\,m \\
        & Limits:  $q_{1}$ = -0.9 to 0.9\,rad, $q_{2}$ = -1.1 to 0.6\,rad, $q_{3}$ = 0.4 to 2.4\,rad\\
    %\midrule
    Actuators & 18 $\times$ Apptronik P170 Orion SEAs \\
    %\midrule
    Controllers & 18 $\times$ Apptronik Axon Whistle 3.2 Controller \\
                & 1 $\times$ Apptronik Medulla Ethercat Network Controller \\
    %\midrule
    Power supply & 14-cell Li-ion battery (58.2\,V, 11800\,mAh) \\
                & External Power (54\,V, 6.0\,A) \\
                 
    %\midrule
    Computers & Autonomy: Intel i7 NUC (32\,GB RAM) \\ 
              & Control: Intel i7 NUC (32\,GB RAM) with RT Kernel \\ 
               
    %\midrule
    Sensors & IMU (Microstrain GX5 - 100\,Hz) \\
   
    \bottomrule
    \end{tabularx}
    \end{center}
    %%\vspace{-0.5cm}
\end{table}

\subsection{Actuator Selection}
Actuator selection is a critical component of designing a new system. Based on the above design requirements and the pursuit of a control strategy with active dynamics, an actuator with accurate force control was required. In order to assess the control requirements for the actuators, an initial kinematic model was developed in simulation using OpenSHC and Gazebo \cite{tam2020openshc}. This simulation provided an initial estimate of the required torques and velocities for each joint. Based on these requirements, quasi-direct-drive (QDD) and series elastic actuators (SEAs) were identified as the most suitable for use in the hexapod system.

\begin{figure}[t!]
    %%\vspace{-0.5cm}
    \centering
    \includegraphics[width=4.5cm]{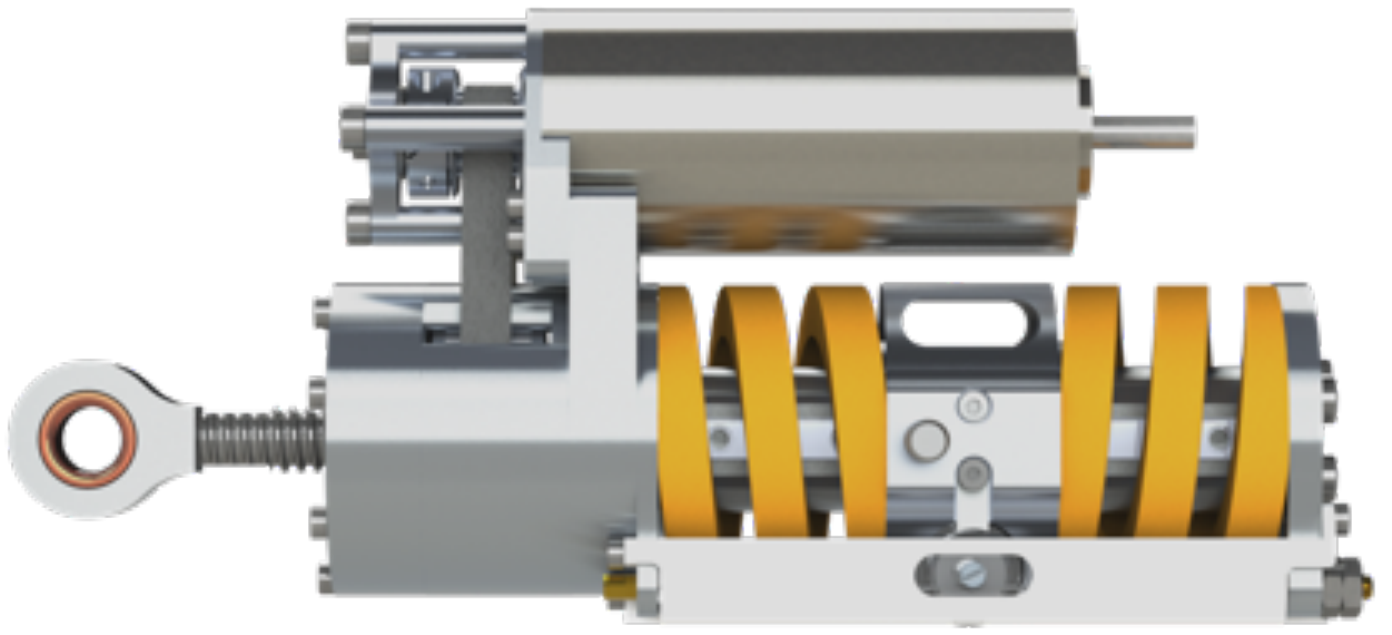}
    \caption{Apptronik P170 Orion linear series elastic actuator~\protect\cite{apptronik}.}	
    \label{fig:actuator}
    \vspace{-0.5cm}
\end{figure}

QDD actuators typically feature a high powered brushed or brushless DC motor mated with a low reduction transmission, typically 5-15:1 reduction. This allows the actuator to achieve the required torque while sensing the motor current for direct force control. Most dynamic legged platforms in this size class employ QDD solutions, including MIT Cheetah\,3~\cite{bledt_mit_2018}, Ghost Vision\,60~\cite{ghostrobotics}, Laikago Pro~\cite{unitree} and Boston Dynamics' Spot~\cite{bd_youtube_2018}. SEAs instead operate by placing a passive spring element in series with the output of the actuator~\cite{pratt_series_1995}. Accurate force control is measured through the spring deflection, allowing for high transmission ratios whilst maintaining force control. Because the spring is an energy absorbing element, SEAs can deal with large impulses passively, such as those that occur during legged locomotion. Examples of robots using SEAs include ANYmal~\cite{hutter_anymal_2016} and the HEBI Robotics hexapods~\cite{travers_dynamical_2016,hebi_daisy}.

\begin{table}[t!]
    \caption{Apptronik Actuator Specifications.}  
    \label{table:apptronik}
    \vspace{-3.5ex}
    \begin{center}
    \begin{tabular}{ccc}
        \toprule
        Peak Force & Continuous Force & Max Linear Speed\\ 
        \midrule
        3200\,N & 847\,N & 0.277\,ms$^{-1}$ \\
        \bottomrule
    \end{tabular}
    \end{center}
        %%\vspace{-0.5cm}

\end{table}

\begin{table}[t!]
    \caption{Theoretical Joint Performance.}  
    \label{table:jointspecs}
    \vspace{-3.5ex}
    \begin{center}
    \begin{tabular}{lrrr}
        \toprule
        Joint & Coxa & Femur & Tibia \\
        \midrule
        Peak Torque (Nm) & 80 & 112 & 80 \\ 
        Continuous Torque (Nm) & 21 & 30 & 21 \\ 
        Max Joint Speed (rads$^{-1}$) & 8 & 11 & 8 \\
        \bottomrule
    \end{tabular}
    \end{center}
    \vspace{-0.5cm}
\end{table}

The Apptronik P170 Orion actuator, based on Paine's work in \cite{paine_high-performance_2014} and shown in Figure~\ref{fig:actuator}, is a high performance linear SEA with specifications shown in Table~\ref{table:apptronik}. When linear SEAs are arranged in a four bar linkage to create a rotary output, the joint space performance can be modified by setting the length of the lever arm in the output linkage. For Bruce, the joints have theoretical performance as shown in Table~\ref{table:jointspecs}.

The drawback of linear SEAs however, is twofold; limited range of motion and non-linear joint performance. However, based on prior knowledge of hexapod design~\cite{elfes_multilegged_2017,bjelonic_weaver:_2018}, a limited range of motion does not hinder walking performance. This is because for the specific morphology chosen, most joint actuation takes place within a limited available workspace. The non-linearity of output torque and speed reduces performance towards the joint limits, however simulation results show the requirements during stable walking are within the performance envelope of the joint.

After all these technical considerations, commercial availability, lead time, cost, and the available development time, the Apptronik P170 Orion actuators were selected for all 18 joints of the robot.

\begin{figure}[b!]
\vspace{-0.5cm}
\centering
\includegraphics[width=5.5cm]{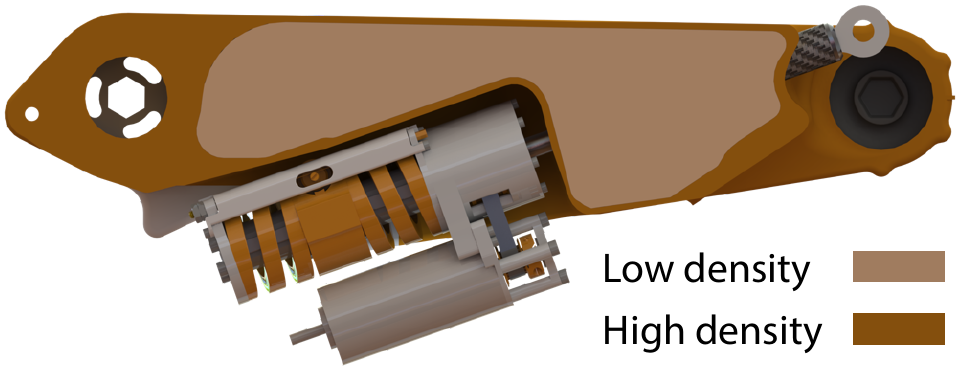}
\caption{Femur cross section showing variable density regions.}
\label{fig:femurcrosssection}
%%\vspace{-0.5cm}
\end{figure}

\subsection{Mechanical and Physical Design}

The core requirement of 3D printing the majority of the robot's structure enabled rapid hardware testing and iterations. The flexibility of 3D printing led to the control of the internal density of components in order to reduce mass while maintaining strength in high stress areas shown in Figure~\ref{fig:femurcrosssection}.  A fast fail paradigm was adopted to iterate on leg revision and actuator configurations. Actuator lever length adjustments were updated during testing to improve the available torque in the femur joint as the expected mass of the system increased as the project matured. Each leg has identically constructed femur and tibia links. The build differs for the coxa links and attachment method to the robot's body due to mirroring two of the coxa configurations in order to package their actuators below the body. Systems related hardware is mounted internally inside the carbon fibre aluminium composite body that provides significant rigidity and torsional stiffness to the robot assembly.  

\begin{figure}[t!]
 %%\vspace{-0.5cm}
    \centering
    \includegraphics[width=6cm]{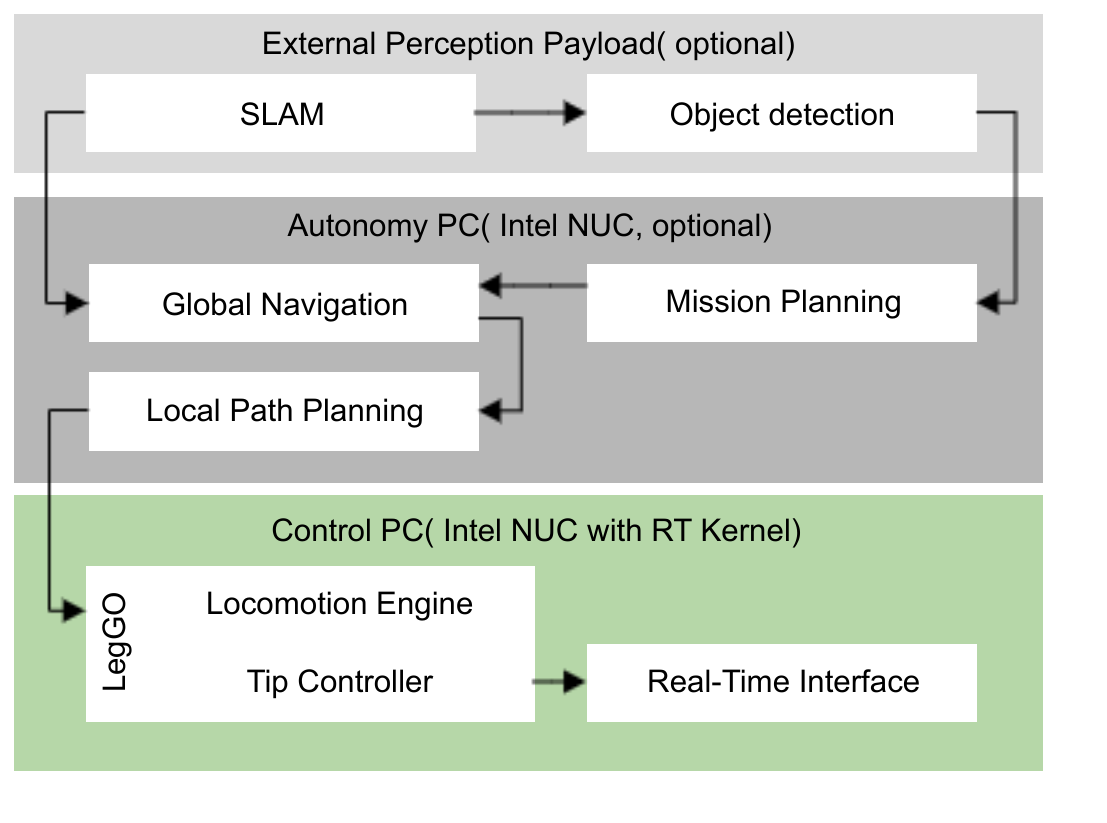}
    \caption{Computer System Architecture}
    \label{fig:autonomyarchitecture}
    \vspace{-0.5cm}
\end{figure}

\begin{figure}[b!]
    \vspace{-0.5cm}
    \centering
    \includegraphics[width=6cm]{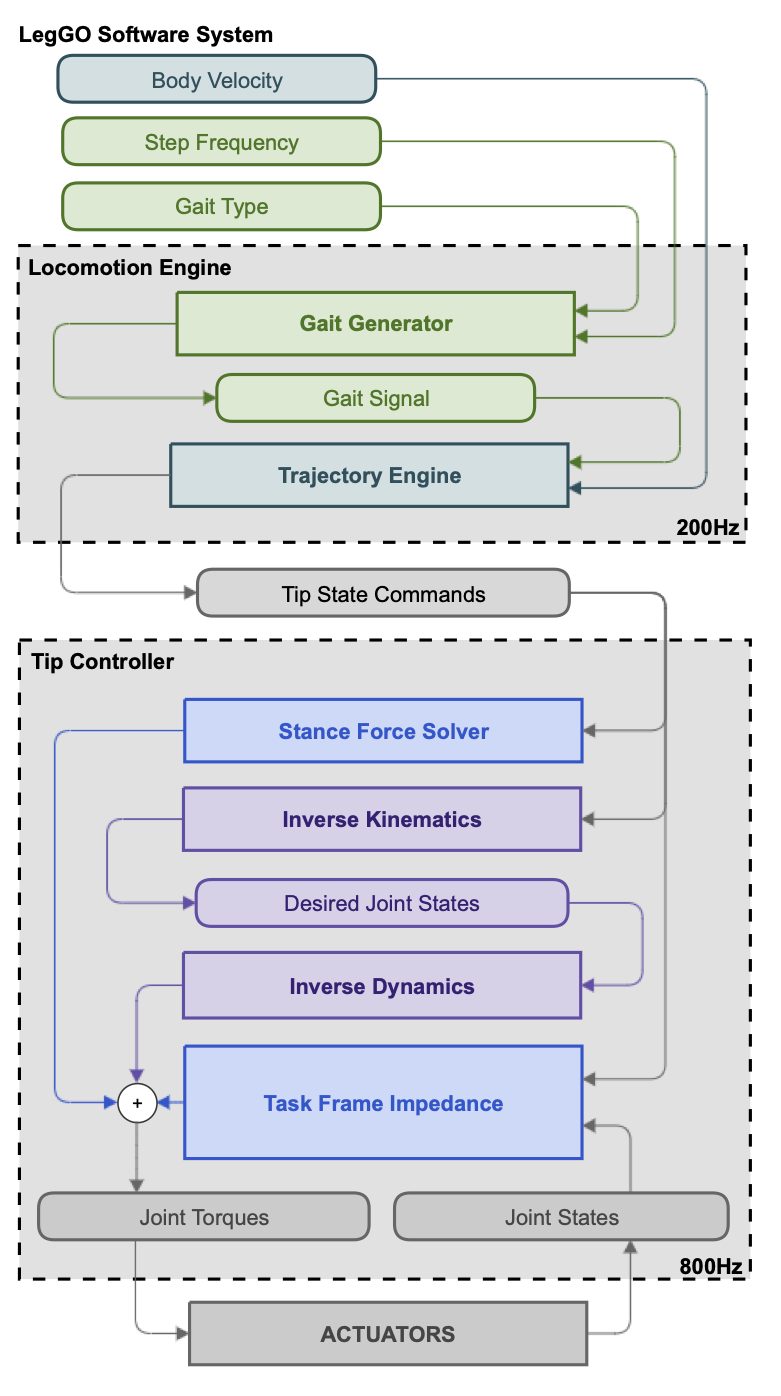}
    \caption{High-level Software Architecture. The rectangular blocks are modules while the curved blocks describe the data types.}
    \label{fig:LeggoArchitecture}
\end{figure}

\subsection{Computer Systems Architecture and Interfacing}

The Apptronik P170 Orion actuators are controlled via Apptronik's Axon Series actuator control boards. These 18 boards are arranged in an EtherCAT loop that communicates with an Intel NUC computer running the control software. The computer is configured with a RT-PREEMPT patched kernel for real-time preempt-ability, allowing the control stack and EtherCAT communications to run at 800\,Hz. To achieve tight time-synchronisation and high data exchange rates required for control, the software modules run inside the Robot Operating System (ROS) nodelet framework with a zero-copy shared memory interface. Apptronik provides an implementation of this framework for synchronising control loops and providing real-time data exchange and logging. Due to the high CPU usage from the real-time control of the hexapod, a second computer is used in parallel for high level autonomy tasks. There is also a second onboard NUC designated for running additional navigation software as required by the mission.

%%%%%%%%%%%%%%%%%%%%%%%%%%%%%%%%%%%%%%%%%%%%%%%%%%%%%%%%%%%%%%%%%%%%%%%%%%

\begin{figure}[t!]
    \centering
    \includegraphics[width=8.5cm]{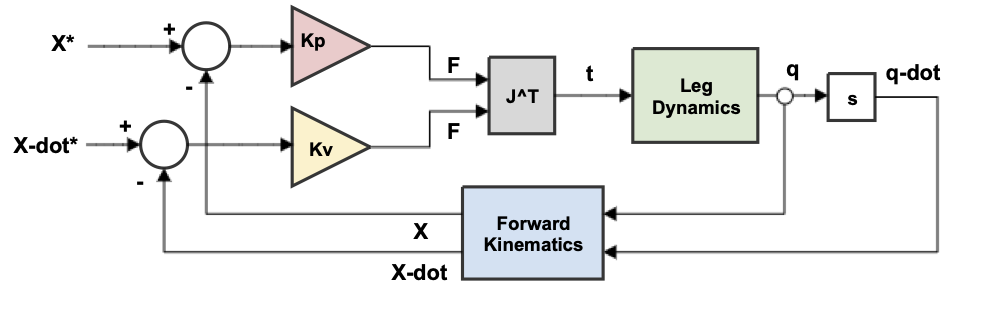}
    \caption{Task Frame Impedance Model}
    \label{fig:jsimpedance}
    \vspace{-0.5cm}

\end{figure}

\subsection{Sensor Payload}
Optional sensor payloads can be mounted at the front of the robot. For the results presented in this paper, a self-contained perception payload performing SLAM was used to provide localisation to the on-board autonomy computer for mission planning. The interface between the autonomy computer and the robot platform is a velocity twist message. Figure~\ref{fig:autonomyarchitecture} shows the computation load distribution of Bruce with the autonomy and control computers. Other sensor payloads up to 5\,kg can be mounted depending on the application.

%%%%%%%%%%%%%%%%%%%%%%%%%%%%%%%%%%%%%%%%%%%%%%%%%%%%%%%%%%%%%%%%%%%%%%%%%

\section{Software Architecture and Control Strategy}
\label{sec:software}

The control strategy of the robot relies on a feed-forward CPG to provide walking patterns, with an underlying task-frame impedance control for each leg to keep the robot stable. During nominal walking, Bruce keeps its CoM within its support polygon over a wide range of speeds, but will become statically unstable as it approaches its fastest design speed of 1.0\,ms$^{-1}$. For lower speed traversal over rough ground (e.g. at 0.3\,ms$^{-1}$), Bruce's primary feedback control mechanism is independent task-frame impedance control of the legs, and tracking the foot position from the CPG. To minimise impedance gains, additional feed-forward inverse dynamic torque setpoints are added to support Bruce's weight and compensate for fast accelerations. This control strategy is implemented through the CSIRO legged locomotion libraries known as `LegGO', a suite of control libraries for multi-legged systems. The libraries are split into two layers. Firstly, the Locomotion Engine takes in target velocities and outputs foot tip trajectories. Secondly, the Tip Controller takes in commanded foot tip states and outputs joint commands. A basic software flow diagram of the LegGO stack is presented in Figure~\ref{fig:LeggoArchitecture}. 

\begin{figure}[!b]
    \vspace{-0.5cm}
    \centering
    \subfigure[]{\includegraphics[width=5cm]{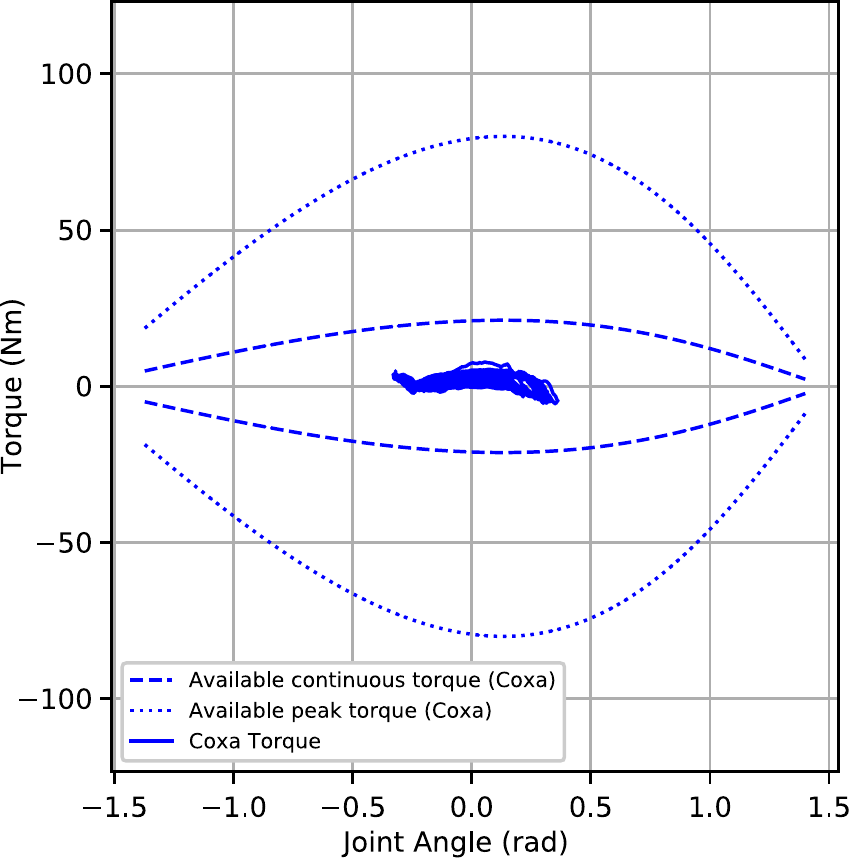}}
    \subfigure[]{\includegraphics[width=5cm]{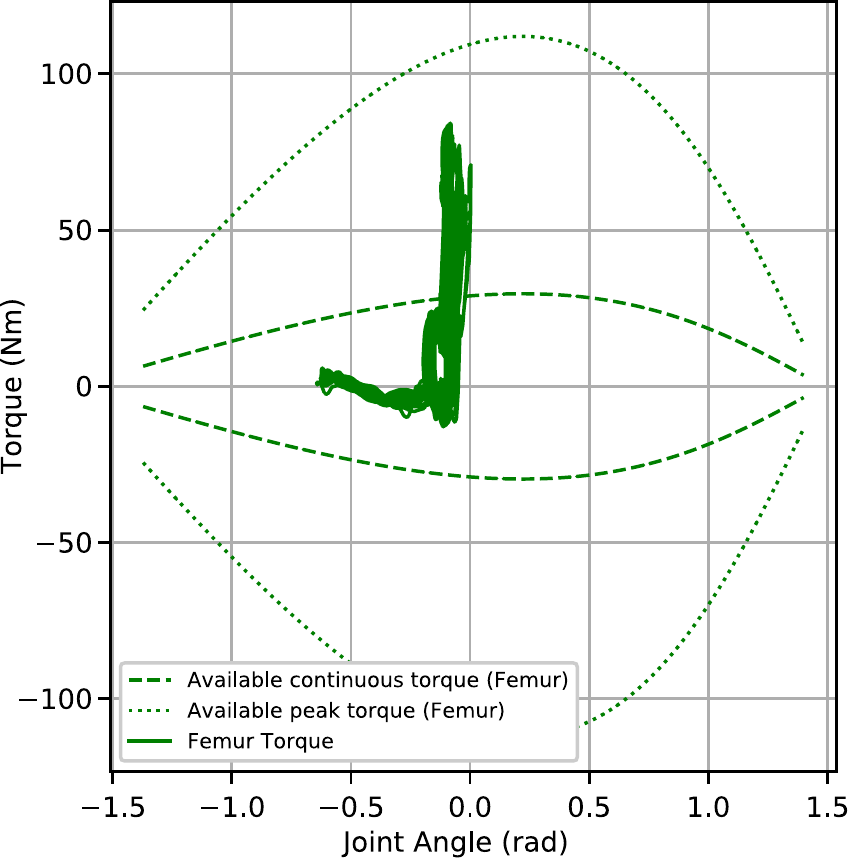}}
    \subfigure[]{\includegraphics[width=5cm]{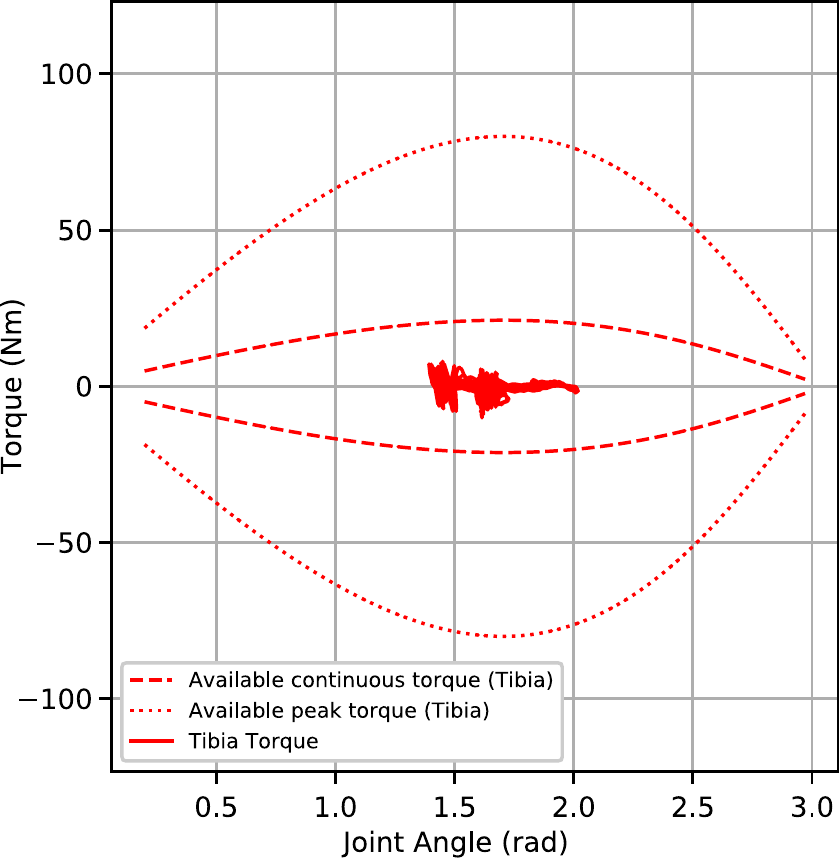}}
    %\vspace{-0.5cm}
    \caption{Calculated torque specifications of latest joint design and experimental joint torque data from flat ground testing for (a) Coxa, (b) Femur and (c) Tibia joints.}
    \label{fig:TorqueRequirementsRealTripod}
    % %\vspace{-0.5cm}
\end{figure}

\subsection{Locomotion Engine}
The Locomotion Engine is an advanced central pattern generator that implements the highest level of control over the locomotion and behaviour of Bruce. For general locomotion, it is split into two separate modules:

\subsubsection{Gait Generator}
This generates a combination of signals for any custom designed gait and a commanded step frequency. The generated gait signal consists of two primary components: the stance and swing signals for the `on-ground' and `in-air' components of the step cycle respectively. The swing signal is further split into three sub-signals: `liftoff', `midswing' and `touchdown' for the associated components of the swing section of the step cycle. These three swing sub-signals and the stance signal feed directly into the four B\'{e}zier curve functions within the Trajectory Engine module.

\subsubsection{Trajectory Engine}
This executes the generation of leg tip trajectory commands, and for each leg consists of four linked 4$^{\text{th}}$ order B\'{e}zier  curve functions for a total of 20 control points per leg trajectory. Each B\'{e}zier  curve respectively handles trajectory generation for the stance, liftoff, mid-swing and touchdown sections of the full step-cycle. The control points of the four B\'{e}zier  curves are calculated to ensure $C^2$ smooth trajectory generation over the entire step cycle whilst achieving the required tip velocity for a commanded body velocity and required touchdown tip positions for commanded foothold locations.

\begin{figure}[t!]
\centering
    \subfigure[]{\label{fig:Bruce_sim} \includegraphics[width=6.5cm]{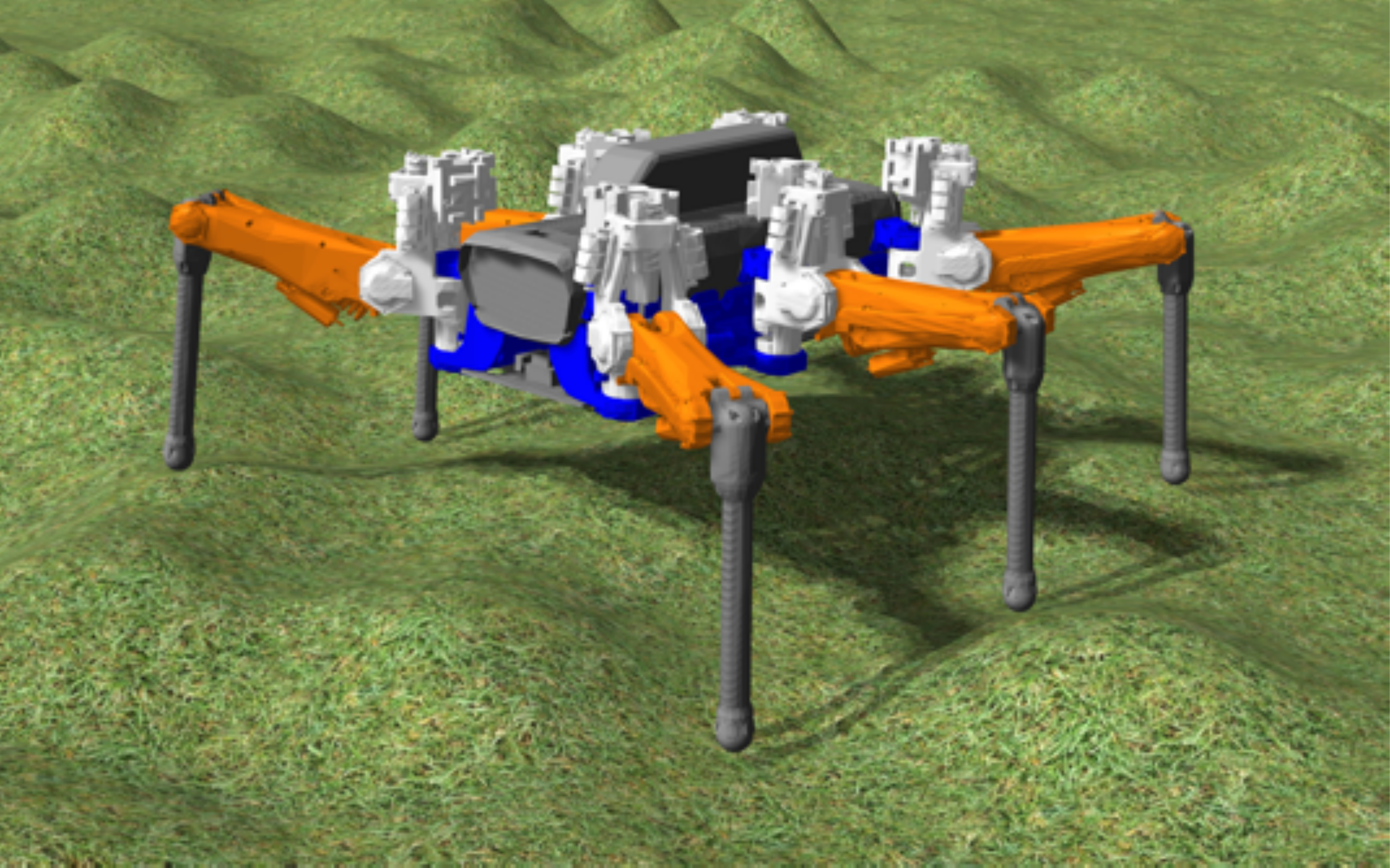}}
    %\vspace{-0.1cm}
    \subfigure[]{\label{fig:Bruce_qual} \includegraphics[width=6.5cm]{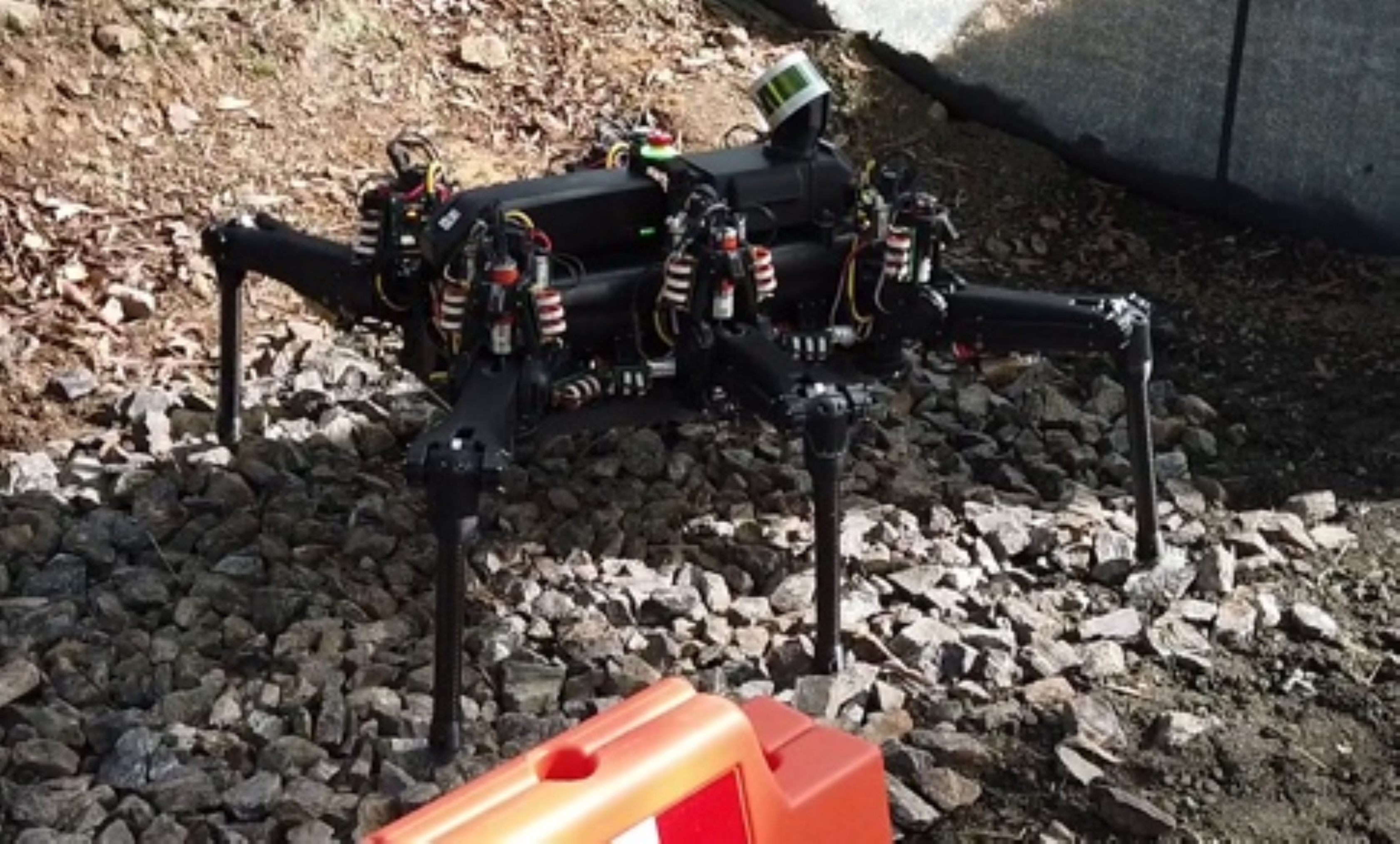}}
    %\vspace{-0.1cm}
\caption{Bruce in (a) Gazebo simulation and (b) rough terrain traversal.}
\vspace{-0.5cm}
\end{figure}

\subsection{Tip Controller}
The Tip Controller performs two functions: tracking of tip trajectories via inverse kinematics, inverse dynamics and impedance control; and management of stance forces for stable walking. The Tip Controller consists of the following modules:

\subsubsection{Stance Force Solver}
This determines the amount of force required at each foot in contact with the ground to maintain a stable stance. This is achieved through assuming a simplified CoM at the centre of the body and then solving the linear system of stance forces using a damped least-squares regression method. These contact forces are converted into feed-forward torques then passed directly to the actuators. 

\subsubsection{Inverse Kinematics}
This (IK) takes in the desired tip positions and velocities from the Trajectory Engine and calculates the desired joint position and velocities to be fed to the Inverse Dynamics (ID) engine. Both the IK and ID engine make use of the Rigid Body Dynamics Library (RBDL) \cite{Felis_2016}, a software library with kinematic and dynamic algorithms for controlling dynamic rigid-body systems.

\subsubsection{Inverse Dynamics}
The ID engine takes in desired joint position, velocity and acceleration to calculate the required joint torques using RBDL algorithms.

\subsubsection{Task Frame Impedance}
This runs at the leg level to control the compliance around the tip at the desired setpoint. A diagram of the model is shown in Figure~\ref{fig:jsimpedance}, where $X$ and $\dot{X}$ are the tip setpoint position and velocity, $q$ and $\dot{q}$ are the joint position and velocity, $F$ is force, $t$ is torque, $J^T$ is the Jacobian transpose and $K_p$, $K_v$ are the impedance model proportional and derivative gains respectively.

%%%%%%%%%%%%%%%%%%%%%%%%%%%%%%%%%%%%%%%%%%%%%%%%%%%%%%%%%%%%%%%%%%%%%%%%%%

%%%%%%%%%%%%%%%%%%%%%%%%%%%%%%%%%%%%%%%%%%%%%%%%%%%%%%%%%%%%%%%%%%%%%%%%%%%%%%%%%%%%%%%%%%%%%%%%%%%
%%%%%%%%%%%%%%%%%%%%%%%%%%%%%%%%%%%%%%%%%%%%%%%%%%%%%%%%%%%%%%%%%%%%%%%%%%%%%%%%%%%%%%%%%%%%%%%%%%%

\begin{figure}[b!]
    %\vspace{-0.5cm}
    \centering
    \subfigure[]{\label{fig:ALTipPositionSimulation}\includegraphics[width=8cm]{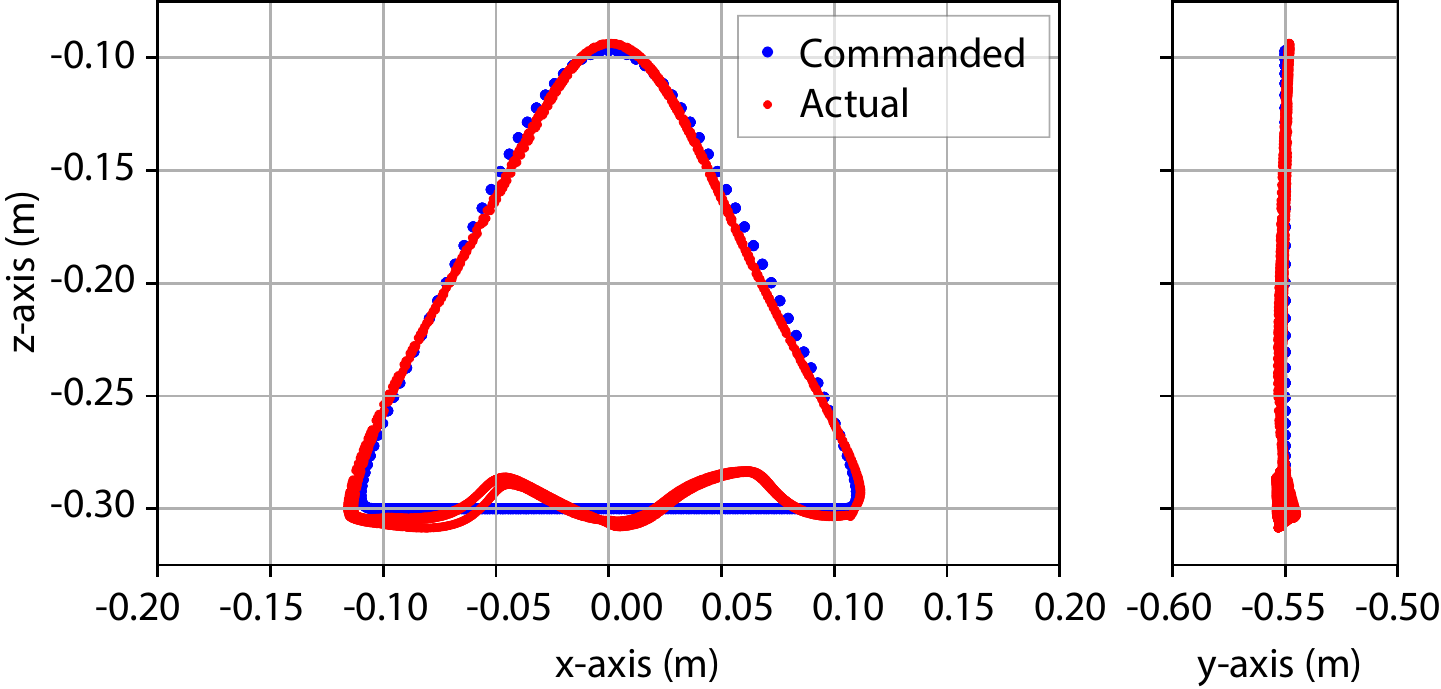}}
    %\vspace{-0.1cm}
    \subfigure[]{\label{fig:BRTipPositionsReal}\includegraphics[width=8cm]{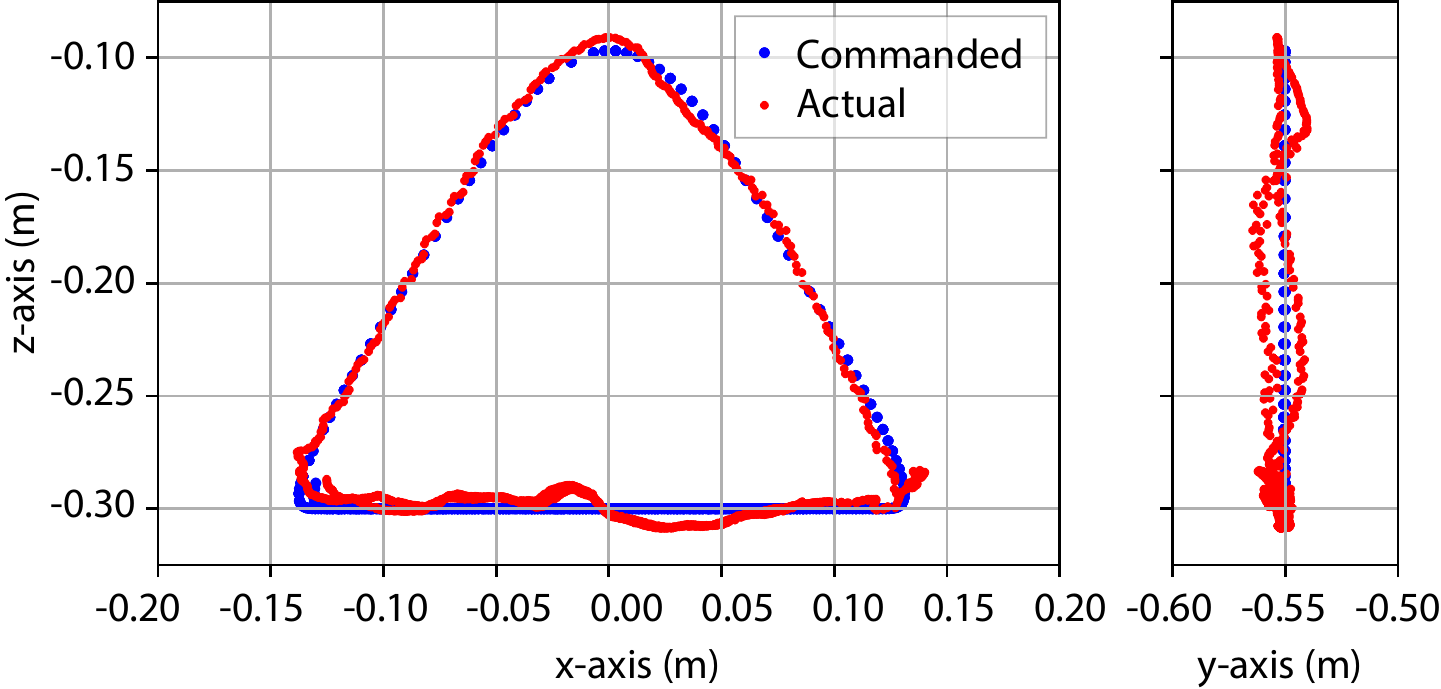}}
    %\vspace{-0.1cm}
    \subfigure[]{\label{fig:BLTipTracking}\includegraphics[width=8cm]{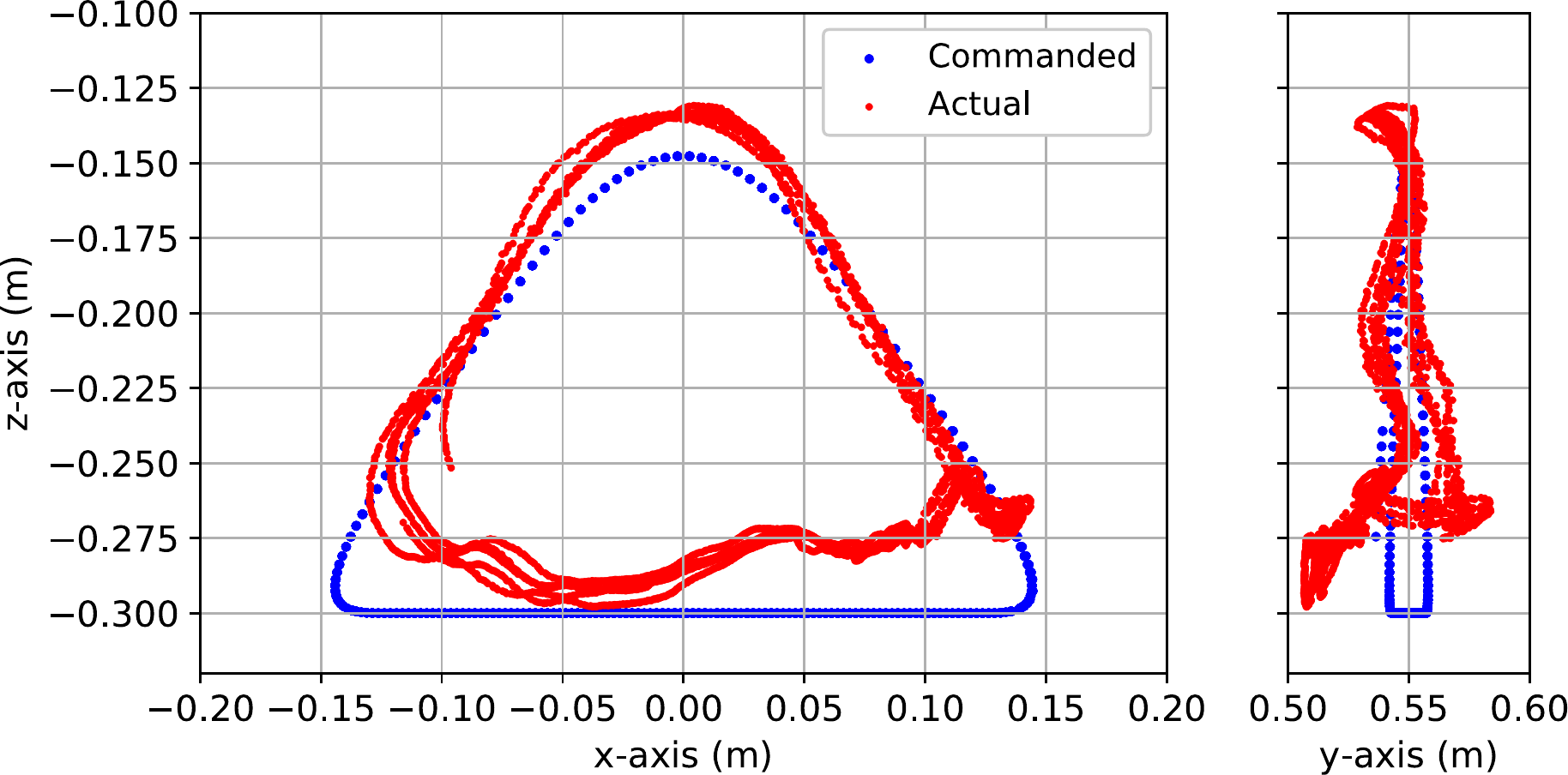}}
    \caption{Commanded vs actual tip position of the middle-right leg (a) in simulation during commanded 0.25\,ms$^{-1}$ velocity with amble gait, (b) during the SubT Tunnel Circuit run with amble gait at 0.75\,Hz step frequency and (c) during tripod walking at 0.5\,ms$^{-1}$ at 1\,Hz step frequency.}
\end{figure}

\section{Performance Evaluation}
\label{sec:evaluation}
\subsection{Simulation Based Testing}
\label{sec:simulation}
Simulation of the robot was developed using the Gazebo environment integrated into ROS Melodic~\cite{koenig_gazebo}. A kinematic simulation was initially used to assess the morphology of the hexapod. For dynamic simulation of the control strategy, the Gazebo ROS Control libraries~\cite{ros_control} was utilised. The simulation model, shown in Figure~\ref{fig:Bruce_sim}, was continually updated to reflect the mass and inertia values of the robot components derived from CAD modelling during periods of heavy design iterations. To validate the actuator requirements, the theoretical achievable torque limits of the joints were overlaid with joint torques generated during a walk cycle of the simulation model, similar to Figure~\ref{fig:TorqueRequirementsRealTripod}. The CPG and impedance control approach described in Section~\ref{sec:software} was tested extensively in simulation before being transferred to hardware. 

\subsection{Field Experiments}
\label{sec:experiments}
The robot was run in a tripod gait for 20\,m at a commanded 0.5\,ms$^{-1}$ in a controlled test environment. These runs were repeated 5 times. Tip tracking performance for this data is presented in Figure~\ref{fig:BLTipTracking}. Qualification for the DARPA SubT Challenge required the robot to navigate a 25\,m course consisting of rock, dirt and grass terrain. Bruce successfully completed the qualification task and is pictured on the rock portion of the course in Figure~\ref{fig:Bruce_qual}.

\begin{figure}[t!]
    %%\vspace{-0.5cm}
    \centering
    \includegraphics[width=8.5cm]{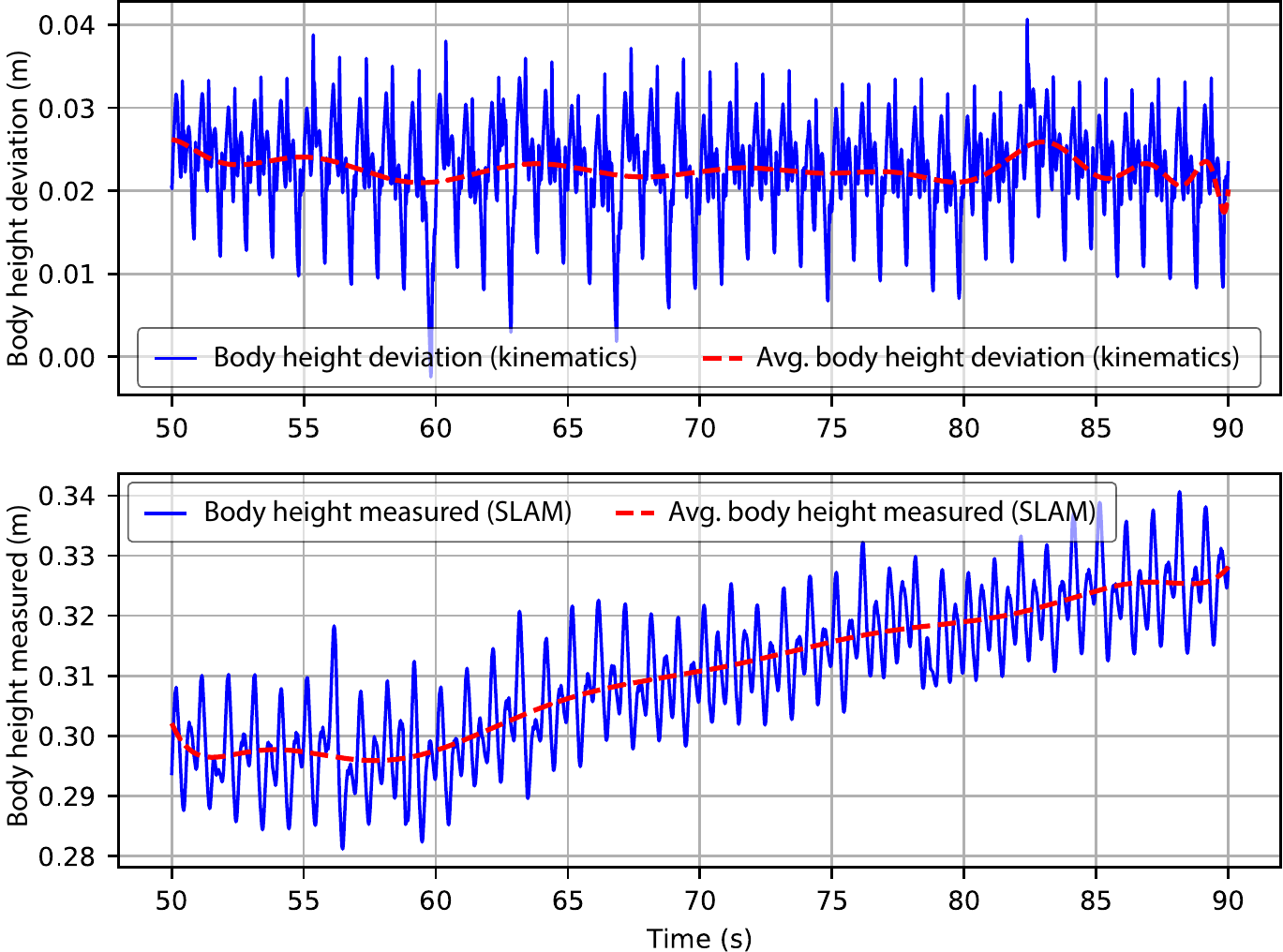}
    %\vspace{-0.5cm}
    \caption{Body height deviation from commanded body height (calculated from leg kinematics) and raw SLAM output of body height during tripod walking.}
    \label{fig:BodyHeightDeviation}
\end{figure}

The robot was also deployed during the Tunnel Circuit event of the DARPA SubT Challenge in Pittsburgh, PA in August 2019. The robot's walking capability in the field was demonstrated by being deployed into the tunnel at a speed of 0.3\,ms$^{-1}$. The tip trajectory tracking data during walking is presented in Figure~\ref{fig:BRTipPositionsReal}. Bruce walked 15\,m into the Experimental Research mine, achieving an average speed of 0.283\,ms$^{-1}$ before the rear left tibia linkage failed.

\begin{figure}[t!]
    %%\vspace{-0.5cm}
    \centering
    \includegraphics[width=8.5cm]{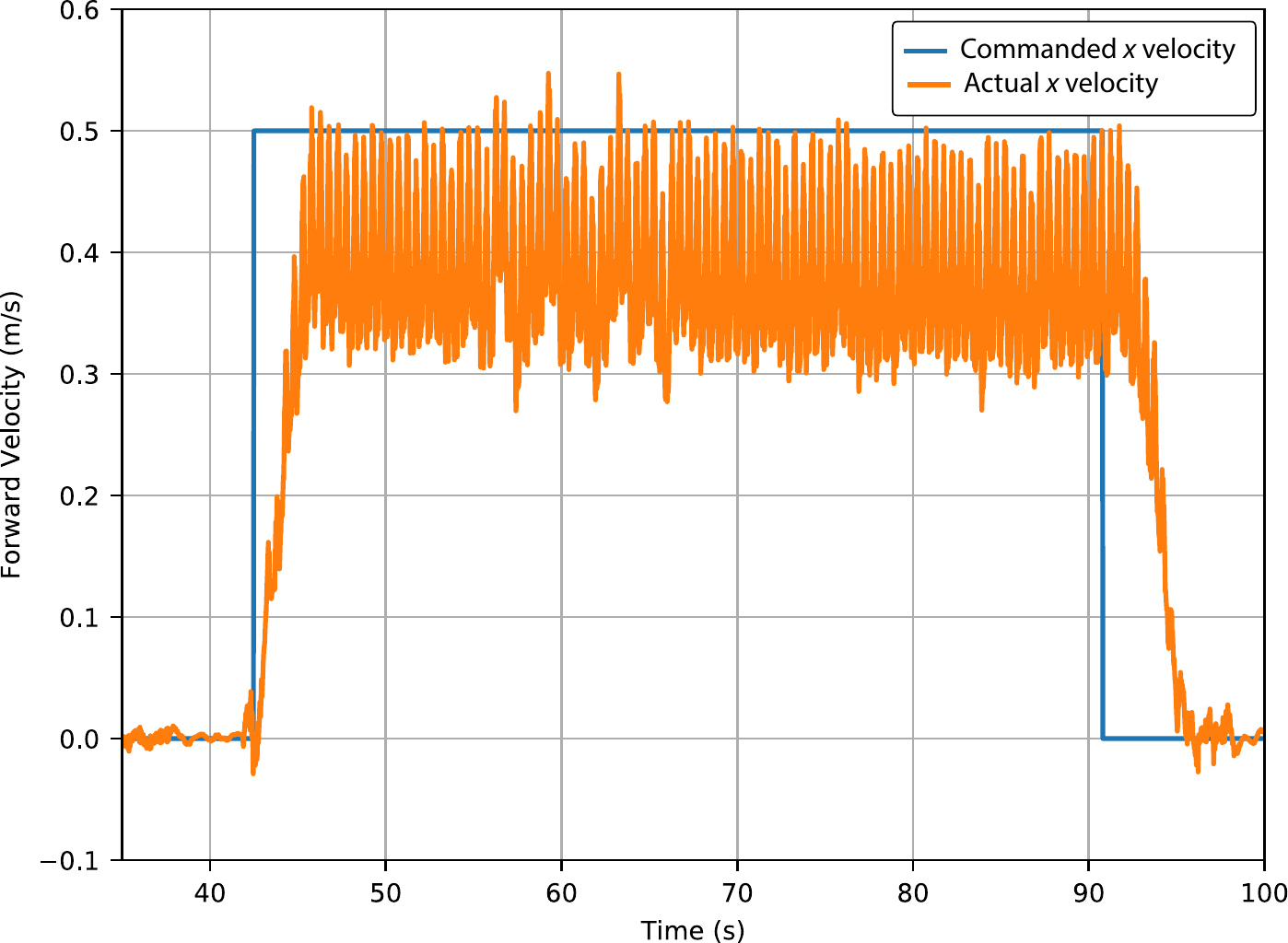}
    %%\vspace{-0.5cm}
    \caption{Velocity tracking along the x-axis during 0.5\,ms$^{-1}$ walking experiment.}
    \label{fig:VelocityTracking}
\end{figure}

\subsection{Analysis}
\label{sec:analysis}
\begin{figure}[b!]
    %%\vspace{-0.5cm}
    \centering
    \includegraphics[height=6cm]{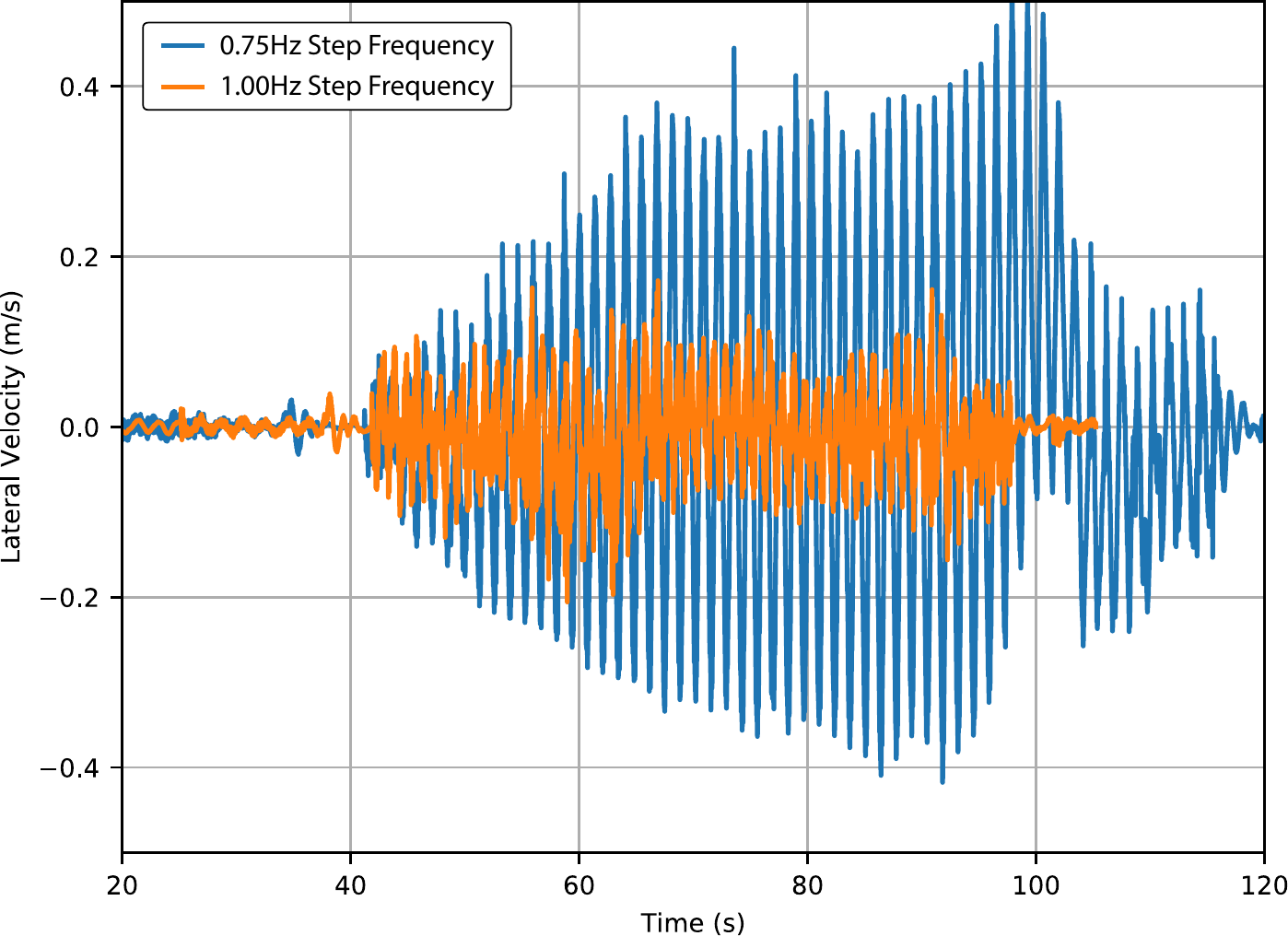}
    %%\vspace{-0.5cm}
    \caption{Lateral velocity comparison between 0.75\,Hz and 1.0\,Hz step frequency.}
    \label{fig:LateralVelComp}
\end{figure}
The experiments demonstrate walking in a tripod gait at 0.5\,ms$^{-1}$. The required torque during regular operation for each of the joints nominally stays below the continuous torque value for the joints, as shown in Figure~\ref{fig:TorqueRequirementsRealTripod}. The exception to this is the femur joint, which peaks above the continuous torque rating during stance. However, this value falls within the peak torque limit for the joint. Thus, the walking data supports the designed performance for each of the joints. Using the on-board SLAM payload and Lidar + IMU EKF pose filter, the pose of the body over time was calculated. The height of the body deviated on average 0.02-0.03\,m peak-to-peak during a gait cycle, shown in Figure~\ref{fig:BodyHeightDeviation}, indicating the feed forward models were adequate at compensating for the robot's mass, even without foot contact detection. The velocity tracking performance of the robot, as determined by the SLAM solution, is shown in Figure~\ref{fig:VelocityTracking}.

Kinematically, the robot is capable of speeds up to 1.4\,ms$^{-1}$ in tripod gait. In tests however, the robot did not demonstrate stable locomotion above 0.5\,ms$^{-1}$. A number of factors prevented the robot from reaching its top speed. Tip trajectory tracking performance was degraded at high step frequencies, due to a relatively low task frame impedance and inaccurate dynamic modelling of the legs. Since the robot is primarily controlled by the CPG, any degradation in tip trajectory tracking reduces system performance. Velocity tracking performance relies upon the task frame impedance, with Figure~\ref{fig:VelocityTracking} showing the robot does not track the desired 0.5\,ms$^{-1}$ velocity and exhibits large oscillations due to the relatively low stiffness of the legs. The desired step frequency from the CPG and its interaction with the leg impedances can excite instabilities in the system. In this case, a 0.75\,Hz step frequency in tripod gait experiences a resonance with the leg impedances and produces large lateral oscillations during walking, as compared with a 1.0\,Hz step frequency, shown in Figure~\ref{fig:LateralVelComp}.

\section{Conclusions and Future Work}
\label{sec:conclusions}
We presented the development and underlying design decisions for Bruce, the CSIRO Dynamic Hexapod Robot designed for dynamic locomotion over rough terrain. Built using linear series elastic actuators, the robot has demonstrated locomotion up to 0.5\,ms$^{-1}$. Using a central pattern generator and task-frame impedance control for individual legs, the control paradigm performs adequately at slow speeds across rough terrain. It is desired to have a low impedance stiffness to gain better disturbance rejection and leg compliance in rough terrain environments. However, when the desired body velocity and step frequency is increased, the tracking performance degrades and the individual task-frame impedance control for each individual leg interacts with each other in an undesirable way. In order to achieve the dual objectives of high leg compliance and stable body locomotion, additional control strategies are required. Future work will focus on providing higher level feedback around the body pose in order to provide more stable walking and improve velocity tracking without the need for higher leg stiffness. To achieve this, it is expected both robust touchdown detection and state estimation will be required.

\section*{Acknowledgements}

This work was funded by the U.S. Government under the DARPA Subterranean Challenge and by Australia's Commonwealth Scientific and Industrial Research Organisation (CSIRO). The authors would like to thank Eranda Tennakoon for feedback on the initial version of the manuscript.

\balance

\bibliographystyle{IEEEtran}
%\bibliography{references}

\end{document}